\documentclass[runningheads]{llncs}
\usepackage[T1]{fontenc}
\usepackage{graphicx}
\usepackage{hyperref}
\usepackage{caption,subcaption}

\usepackage{wrapfig}

\usepackage[dvipsnames]{xcolor}
\definecolor{blue}{rgb}{0.0, 0.5, 0.8}
\definecolor{gray}{rgb}{0.5, 0.5, 0.5}
\definecolor{red}{rgb}{0.9, 0.1, 0.1}
\definecolor{green}{rgb}{0.1, 0.7, 0.1}
\definecolor{cyan}{rgb}{0.1, 0.7, 0.7}

\usepackage{tikz}

\newcommand{\operator}{k}
\newcommand{\inputSig}{R}

\newcommand{\mask}{M}
\newcommand{\outputSig}{B}
\newcommand{\homMat}{H}

\newcommand{\task}{\mathcal{O}}

\newcommand{\dataset}{\mathcal{D}}

\newcommand{\noise}{\eta}
\newcommand{\SDnoise}{\sigma}

\newcommand{\Param}{\ensuremath{\Theta}}
\newcommand{\Net}{\ensuremath{\mathcal{F}}}
\newcommand{\loss}{\ensuremath{\mathcal{L}}}

\newcommand{\disMeasure}{L}

\begin{document}

\title{Multi-task Video Enhancement for \\Dental Interventions\thanks{This work was supported in part by The National Centre for Research and Development, Poland, under grant agreement POIR.01.01.01-00-0076/19.}}
% \title{Multi-task Curriculum Learning for Dental Video Deblurring, Stabilization and Segmentation\thanks{Supported by organization x.}}


\author{Efklidis Katsaros\inst{1}\orcidID{0000-0002-0261-9187} \and
Piotr K. Ostrowski\inst{1}\orcidID{0000-0003-1663-5839} \and
Krzysztof Włódarczak\inst{1}\orcidID{0000-0003-0999-5887} \and
Emilia Lewandowska\inst{1}\orcidID{0000-0001-5556-6640} \and
Jacek Ruminski\inst{1}\orcidID{0000-0003-2266-0088} \and
Damian Siupka-Mróz\inst{4} \and
Łukasz Lassmann \inst{3,4}\orcidID{0000-0002-4974-2625} \and
Anna Jezierska\inst{2,1}\orcidID{0000-0001-8235-7641} \and
Daniel Wesierski\inst{1}\orcidID{0000-0001-7093-8764}}

\authorrunning{E. Katsaros et al.}

\institute{Gdańsk University of Technology \\ \email{daniel.wesierski@pg.edu.pl} \and
Systems Research Institute, Polish Academy of Sciences \and
Dental Sense Medicover \and
Master Level Technologies}

\maketitle              % typeset the header of the contribution
\begin{abstract}

A microcamera firmly attached to a dental handpiece allows dentists to continuously monitor the progress of conservative dental procedures. Video enhancement in video-assisted dental interventions alleviates low-light, noise, blur, and camera handshakes that collectively degrade visual comfort. To this end, we introduce a novel deep network for multi-task video enhancement that enables macro-visualization of dental scenes. In particular, the proposed network jointly leverages video restoration and temporal alignment in a multi-scale manner for effective video enhancement. Our experiments on videos of natural teeth in phantom scenes demonstrate that the proposed network achieves state-of-the-art results in multiple tasks with near real-time processing. We release \textit{Vident-lab} at \href{https://doi.org/10.34808/1jby-ay90}{https://doi.org/10.34808/1jby-ay90}, the first dataset of dental videos with multi-task labels to facilitate further research in relevant video processing applications.

\end{abstract}

\section{Introduction}
%Problem -> applications -> stl parallel -> no, mtl -> related works -> our contributions

\begin{figure}[!htbp]
    \centering
    \centerline{\includegraphics[width=0.95\textwidth]{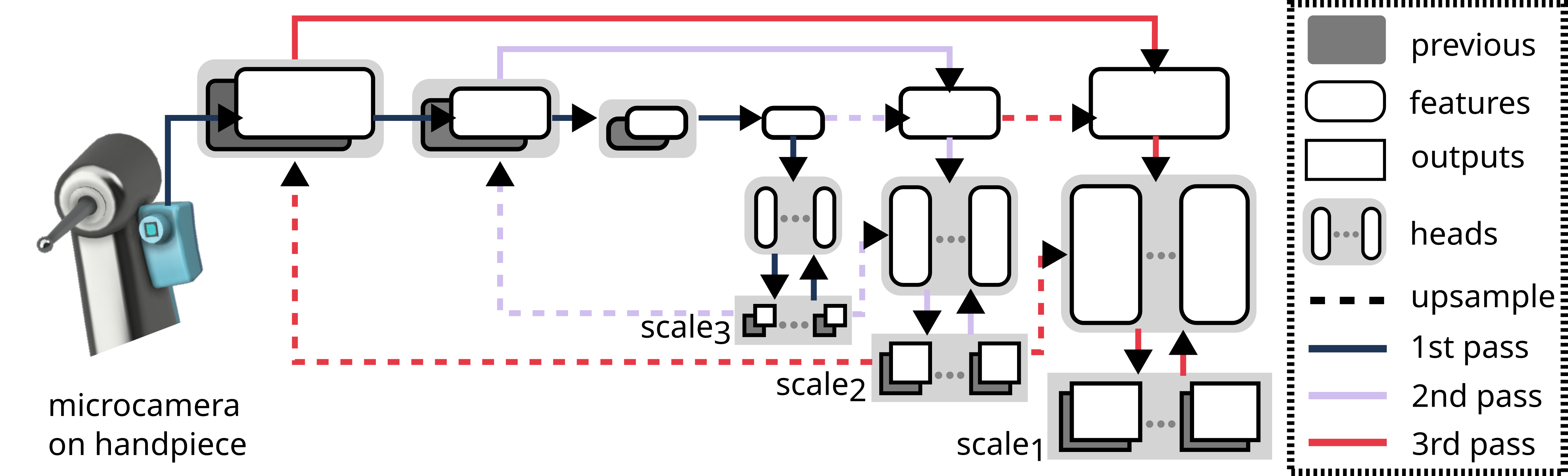}}
    \caption{Multi-task enhancement of videos from dental microcameras. We propose \textbf{MOST-Net}, a multi-output, multi-scale, multi-task network that propagates task outputs within and across scales in the encoder and decoder to fuse spatio-temporal information and leverage task interactions.
    %The acquired low-quality video requires algorithmic enhancement to offer a continuous, near-range, and high-quality display of the operated tooth.
    }
    \label{fig:flowchart}
\end{figure}

% Why our microcamera and our tasks
Computer-aided dental intervention is an emerging field \cite{zhu2020tooth,kuhnisch2021caries,rashid2022hybrid}. In contemporary clinical practice, dentists use various instruments to view the teeth better for decreased work time and increased quality of conservative dental procedures \cite{low2018magnification}. A close and continuous view of the operated tooth enables a more effective and safer dental bur maneuver within the tooth to remove caries and limit the risk of exposing pulp tissue to infection. A microcamera in an adapter firmly attached to a dental handpiece near the dental bur allows dentists to inspect the operated tooth during drilling closely and uninterruptedly in a display. However, the necessary miniaturization of vision sensors and optics introduces artifacts. Macro-view translates the slight motion of the bur to its more significant image displacement. The continuous camera shakes increase eye fatigue and blur. Handpiece vibrations, rapid light changes, splashing water and saliva further complicate imaging of intra-oral scenes. This study is the first to address the effectively compromised quality of videos of phantom scenes with an algorithmic solution to integrate cost-effective microcameras into digital dental workflows.

We propose a new multi-task, decoder-focused architecture \cite{vandenhende2021multi} for video processing and apply it to video enhancement of dental scenes (Fig. \ref{fig:flowchart}). The proposed network has multiple heads at each scale level. Provided that task-specific outputs amend themselves to scaling, the network propagates the outputs bottom-up, from the lowest to the highest scale level. It thus enables task synergy by loop-like modeling of task interactions in the encoder and decoder across scales. UberNet \cite{kokkinos2017ubernet} and cross-stitch networks \cite{misra2016cross} are encoder-focused architectures that propagate task outputs across scales in the encoder. Multi-modal distillation in PAD-Net \cite{xu2018pad} and PAP-Net \cite{zhang2019pattern} are decoder-focused networks that fuse outputs of task heads to make the final dense predictions but only at a single scale. MTI-Net \cite{vandenhende2020mti}, which is most similar to our architecture, extends the decoder fusion by propagating task-specific features bottom-up across multiple scales through the encoder.
% (, convincingly arguing task interactions can vary with the size of receptive fields.
Instead of propagating the \textit{task features} in scale-specific distillation modules across scales to the \textit{encoder}, our network simultaneously propagates \textit{task outputs} to the \textit{encoder} and to the task heads in the \textit{decoder}. Furthermore, the networks make dense task prediction in static images while we extend our network to videos. 

%address the tasks of video deblurring, denoising, color correction, binary segmentation, and motion estimation in near real-time.

We instantiate the proposed model to jointly solve low-, mid-, and high-level video tasks that enhance intra-oral scene footage. 
%in the presence of low-light, sensor noise, motion blur, and camera handshakes. 
In particular, the model formulates color mapping~\cite{zhang2020rt}, denoising, and deblurring \cite{wang2019edvr,zhong2020efficient,katsaros2021concurrent} as a single dense prediction task and leverages, as auxiliary tasks, homography estimation \cite{le2020deep} for video stabilization \cite{bradley2021cinematic} and teeth segmentation~\cite{chen2018encoder,zhou2018unet++} to re-initialize video stabilization. Task-features interacted in the two-branch decoder in \cite{jung2021multi} at a single scale for dense motion and blur prediction in dynamic scenes. We are the first to jointly address the tasks of color mapping, denoising, deblurring, motion estimation, and segmentation. 
% We demonstrate that our near real-time network achieves state-of-the-art results in these tasks in comparison to corresponding single-task networks on videos with natural teeth in phantom scenes.
We demonstrate that our near real-time network achieves state-of-the-art results in multiple tasks on videos with natural teeth in phantom scenes.

%In near-range imaging, users are sensitive to artificial deformations of rigid regions such as a throbbing gum and teeth. In contrast to deep mesh-based video stabilization \cite{wang2018deep,pwstablenet} that struggled to enforce object rigidity, we restrict cameras to follow a global homography trajectory to better preserve the rigid appearance of the teeth. 

Our contributions are: (i) a novel application of a microcamera in computer-aided dental intervention for continuous tooth macro-visualization during drilling, (ii) a new, asymmetrically annotated dataset of natural teeth in phantom scenes with pairs of frames of compromised and good quality using a beam splitter, (iii) a novel deep network for video processing that propagates task outputs to encoder and decoder across multiple scales to model task interactions, and (iv) demonstration that an instantiated model effectively addresses multi-task video enhancement in our application by matching and surpassing state-of-the-art results of single task networks in near real-time.

\section{Proposed method}

Video enhancement tasks are interdependent, e.g. aligning video frames assists deblurring \cite{zhong2020efficient,zhou2019stfan,katsaros2021concurrent,wang2019edvr} while denoising and deblurring expose image features that facilitate motion estimation \cite{le2020deep}. We describe MOST-Net (multi-output, multi-scale, multi-task), a network that models and exploits %such
task interactions across scale levels of the encoder and decoder. We assume the network yields $T$ task outputs at each scale $\{{\task}_{i}^{s}\}_{i,s=1}^{T,S}$, where $s=1$ denotes the original image resolution. 
Task outputs are propagated innerscale but also upsampled from the lower scale and propagated to the encoder layers and the task-specific branches of the decoder at higher scales.
We require the following task-specific relation:   
\begin{equation}
u_{i}(\task_{i}^{s+1}) \approx \task_{i}^s,
\label{eq:output_scale}
\end{equation}
where $u_{i}$ denotes some operator, for instance, the upsampling operator for segmentation or the scaling operator for homography estimation. 

%For example, homography estimation task can be scaled however image classification task yields semantic labels that are unscalable %\pko{(I think that s in $u(\task_{i}^{s+1};s)$ is not necessary. I think that using term 'upsampling' for scaling up homography is not correct, I'd use more general phrase 'scale up' instead of narrower 'upsample' but I don't know if it is correct)}.

\begin{figure}[]
    \centering
    \centerline{\includegraphics[width=0.99\textwidth]{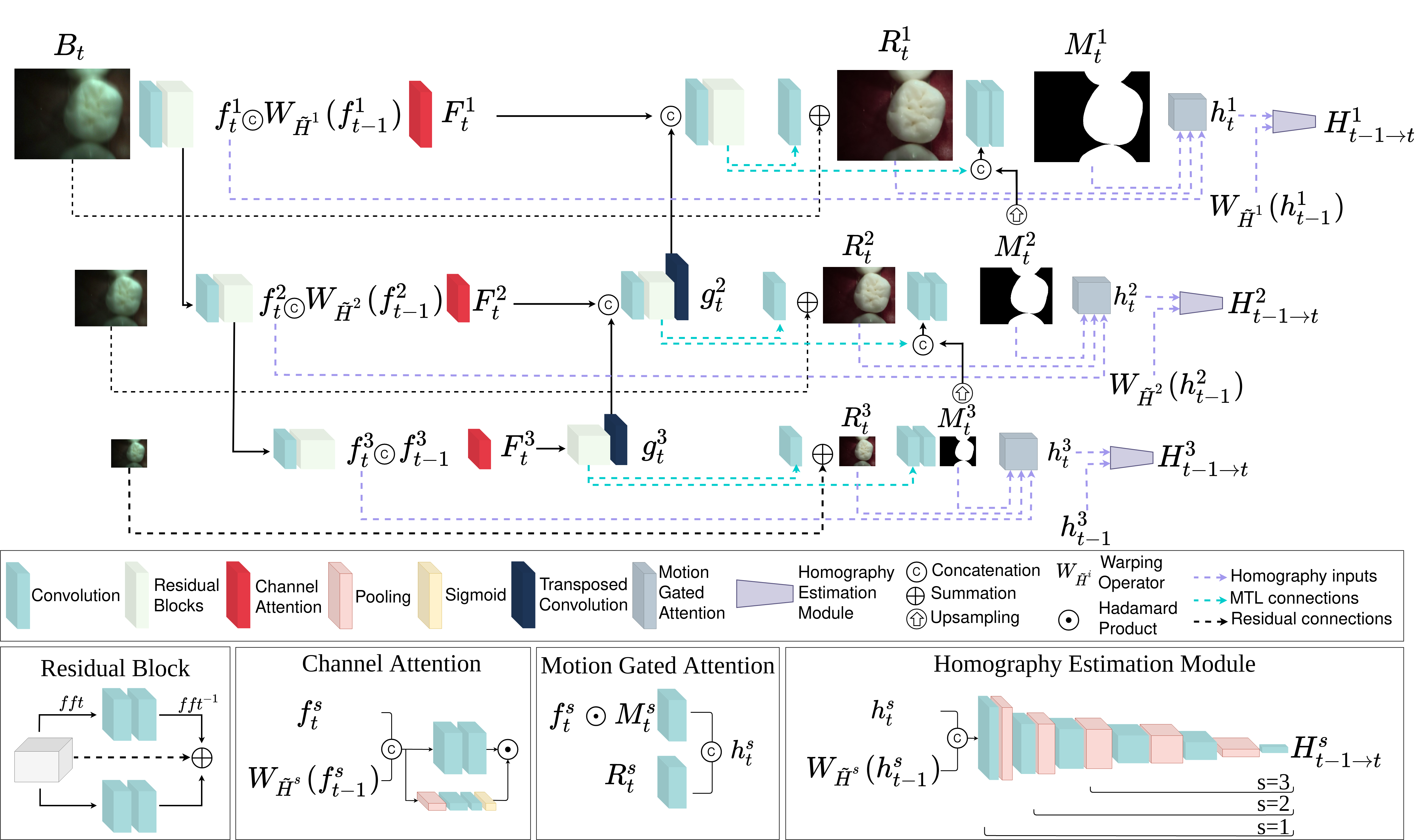}}
    \caption{Our MOST-Net instantiation addresses three tasks for video enhancement: video restoration, teeth segmentation, and homography estimation.}
    \label{fig:arch}
\end{figure}

{\bf{Problem Statement.}} We instantiate MOST-Net to address the video enhancement tasks in dental interventions. In this setting, $T=3$ and $\task_1$, $\task_2$, $\task_3$ denote the outputs for video restoration, segmentation and homography estimation. A video stream generates observations $\{\outputSig_{t-z}\}_{p=0}^P$, where $t$ is the time index and $P>0$ is a scalar value referring to the number of past frames. The problem is to estimate a clean frame, a binary teeth segmentation mask and approximate the inter-frame motion by a homography matrix, denoted by the triplet ${\task}_{1,1}^{3,S} = \{\inputSig_{t}^s$,  $\mask_{t}^s,$ $\homMat_{t-1 \rightarrow t}^s,\}_{s=1}^S$.
Let $x$ correspond to pixel location. Given per-pixel blur kernels $\operator_{x,t}$ of size $K$, the degraded image at $s=1$ is generated as:
\begin{equation}
    \forall_x \; \forall_t \;  \outputSig_{x,t} = \SDnoise \langle (\inputSig_{x,t})^- \operator_{x,t} \rangle + \noise,
\end{equation}
where $\noise$ and $\SDnoise$ denote additive and signal dependent noise, respectively, while $(\inputSig_{x,t})^-$ is a window of size $K$ around pixel $x$ in image $\inputSig_{t}^1$. 
%Such a space and time variant blur may account for motion blur or defocus blur. 
Next we assume multiple independently moving objects present in the considered scenes, while our task is to estimate only the motion related to the object of interest (i.e. teeth), which is present in the region indicated by non-zero values of mask $M$:
\begin{equation}
  \forall_{t} \; \; \forall_{x} \qquad \qquad x_{t} = H_{t-1 \rightarrow t} x_{t-1} \quad s.t. \quad M_{t}(x) = 1
\end{equation}

\textbf{Training.} In our setting, all tasks share training data in dataset $\dataset = \{\{\outputSig \}_j, \{{\task}_{i}^{s}\}_j\}_{i,s,j=1}^{T,S,N}$, where $\{\task_{i}^s\}_j$ is a label related to task $i$ at scale $s$ for the $j$-th training sample $\{\outputSig\}_j$, while $N$ denotes number of samples in training data. In the context of deep learning, the optimal set of parameters $\Param$ for some network $\Net_{\Param}$ is derived by minimizing a penalization criterion: 
\begin{equation}
      \loss(\Param)=\sum_j^N \sum_i^T \sum_s^S \lambda_i \disMeasure_i \left(\{\task_{i}^s\}_j, \{\hat{\task}_i^s (\Param)\}_j \right), 
\label{eq:loss}
\end{equation}
where $\lambda_i$ is a scalar weighting value, $\hat{\task}_i^s (\Param)$ is an estimate of $\task_{i}^s$ for $j$-th sample in $\dataset$, and $\disMeasure_i$ is a distance measure. In this study, we use the Charbonnier loss~\cite{charbonnier1994two} as $\disMeasure_1$, the binary cross-entropy~\cite{murphy2012machine} as $\disMeasure_2$ and the Mean Average Corner Error (MACE)~\cite{detone2016deep} as $\disMeasure_3$.

{\bf{Encoders.}} At each time step, MOST-Net extracts features $f_{t-1}^s$, $f_t^s$ from two input frames $B_{t-1}$ and $B_t$ independently at three scales. To effectuate the U-shaped~\cite{ronneberger2015u} downsampling, features are extracted via $3 \times 3$ convolutions with strides of $1,2,2$ for $s=1,2,3$ followed by ReLU activations and 5 residual blocks~\cite{cho2021rethinking} at each scale. The residual connections are augmented with an additional branch of convolutions in the Fast Fourier domain, as in \cite{mao2021deep}. The output channel dimension for features $f_t^s$ is ${2^{s+4}}$. At each scale, features $f_{t}^s$ and $W_{\tilde H_s}(f_{t-1}^s)$ are concatenated and a channel attention mechanism follows \cite{zhong2020efficient} to fuse them into $F_t^s$. MOST-Net uses homography outputs from lower scales to warp encoder features from the previous time step as $W_{\tilde{H}} (f_{t-1}^s)$. Here, $\tilde{H}^s$ is an upscaled version of $H^{s+1}$ for higher scales and the identity matrix for $s=3$.

{\bf{Decoders.}} The attended encoder features $F_t^s$ are passed onto the expanding blocks scale-wisely via the skipping connections. At the lower scale ($s=3$), the attended features $F_{t}^3$ are directly passed on a stack of two residual blocks with $128$ output channels. Thereafter, transposed convolutions with strides of $2$ are used twice to recover the resolution scale. At higher scales ($s<3$), features $F_{t}^s$ are first concatenated with the upsampled decoder features $g_{t}^{s-1}$ and convolved by $3 \times 3$ kernels to  halve the number of channels. Subsequently, they are propagated onto two residual blocks with 64 and 32 output channels each. The residual block outputs constitute scale-specific shared backbones. Lightweight task-specific branches follow to estimate the dense outputs. Specifically, one $3 \times 3$  convolution estimates $R_t^s$ and two $3 \times 3$ convolutions, separated by ReLU, yield $M_t^s$ at each scale. Fig.~\ref{fig:arch} shows MOST-Net enables refinement of lower scale segmentations by upsampling and inputting them at the task-specific branches of higher scales.

At each scale, homography estimation modules estimate $4$ offsets, related 1-1 to homographies via the Direct Linear Transformation (DLT) as in~\cite{detone2016deep,le2020deep}. The motion gated attention modules multiply features $f_t^s$ with segmentations $M_t^s$ to filter out context irrelevant to the motion of the teeth. The channel dimensionality is then halved by a $3\times3$ convolution while a second one extracts features from the restored output $R_t^s$. The concatenation of the two streams forms features $h_t^s$. At each scale, $h_t^s$ and $W_{\tilde{H}^s}(h_{t-1}^s)$, are employed to predict the offsets with shallow downstream networks. Predicted offsets at lower scales are transformed back to homographies and cascaded bottom-up~\cite{le2020deep} to refine the higher scale ones. Similarly to~\cite{detone2016deep}, we use blocks of $3\times3$ convolutions coupled with ReLU, batch normalization and max-pooling to reduce the spatial size of the features. Before the regression layer, a $0.2$ dropout is applied. For $s=1$, the convolution output channels are 64, 128, 256, 256 and 256. For s=2,3 the network depth is cropped from the second and third layers onwards respectively.

\section{Results and Discussion}

%The proposed pipeline enjoys the merits of a) the spatial alignment issue is circumvented via deep learning of the CM function that is directly applied on the denoised frames b) a realistic motion blur model with respect to the individual object motion trajectories is utilized c) The N2N denoising configuration retains the real camera noise. The dataset is publicly released under the link. 

%\ek{(What is the notation finally here $R$, or $R_t$)}
%$\dataset = \{ \{B\}_j, \{R^s\}_j, \{M^s\}_j, \{H_{t-1,t}^s\}_j \}_{s,j=1}^{S,N}$ 
%\pko{(I do not understand reason of using 't' for $B_t, R_t, M_t$)} 

\begin{figure}[]
    \centering
    \includegraphics[width=\textwidth]{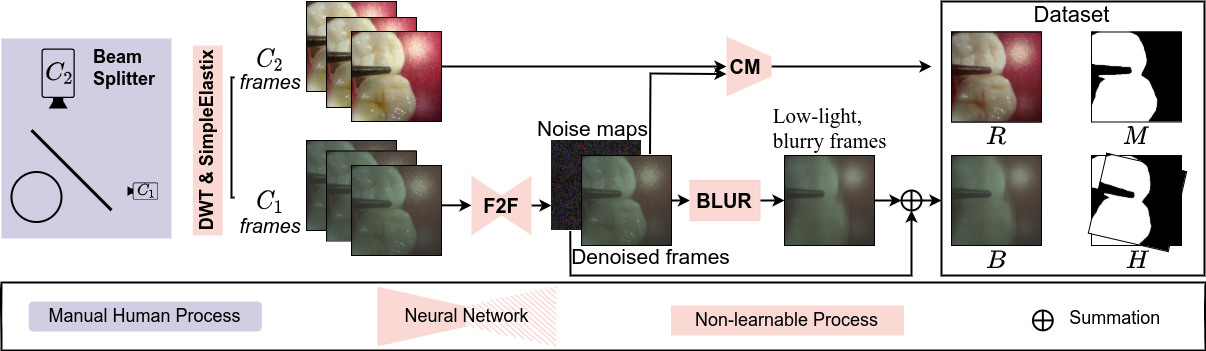}
    \caption{A flowchart of dataset preparation.}
    %Our video recording setup uses a beam splitter. Spatial alignment between the two sensors is circumvented by learning a color mapping.}
    \label{fig:data}
\end{figure}

{\textbf{Dataset.}} We describe the generation of \textit{Vident-lab} dataset $\dataset$ with frames $\outputSig$ and labels $R$, $M$ and $H$ for training, validation, and testing (Tab. \ref{tab:db}). We generate the labels at full resolution as illustrated in Fig. \ref{fig:data}. The lower scale labels $\task_{i}^s$ are obtained from the inverse of Eq.~\ref{eq:output_scale}, i.e. downsampling for $R$, $M$ and downscaling for $H$. The  dataset is publicly available at \href{https://doi.org/10.34808/1jby-ay90}{https://doi.org/10.34808/1jby-ay90}.

%\pko{(Why B is not mentioned here?)}. 
%multi-task video enhancement network
%We create (i) synthetically corrupted video frames $B_t$ and clean, ground truth frames $R_t$, (ii) manually and algorithmically obtained segmentation masks $M_t$, and (iii) homography matrices between pairs of consecutive frames $H_{t-1,t}$ 

% \aj{We should explain here also how we obtains the labels at scales 2 and 3}
% \pko{Proposition for explanation for scale 2 and 3: Below method of obtaining labels for the highest scale is described. To get label for task i and scale $s>1$  we use operation inverse to u_i:
% \begin{equation}
% %\uparrow \task_{i}^{s+1} = u(\task_{i}^{s+1};s) \approx \task_{i}^s,
% \task_{i}^s = u_{i}^{-1}(\task_{i}^{s-1}),
% \end{equation}
% }

\textbf{Data acquisition.} A miniaturized camera $C_1$ has inferior quality to intraoral cameras $C_2$, which have larger sensors and optics. Our task is to teach $C_1$ to image the scene equally well as $C_2$. Both cameras, which are firmly coupled through a $50/50$ beam splitter, acquire videos of the same dental scene. Dynamic time warping (DTW) synchronizes the videos
% as tracklets of $C_1$ and $C_2$ are the same, 
and then \textit{SimpleElastix} \cite{marstal2016simpleelastix} registers the corresponding $320\times416$ frames.

\begin{wraptable}{r}{4cm}
\centering
\begin{tabular}{ c|c|c|c  }
data      & train & val  & test   \\ \hline \hline
videos    & 300   & 29   & 80     \\
frames    & 60K   & 5.6K & 15.5K  \\
segm(h)   & 300   & 116  & 320    \\
segm(n)   & 59.7K & 5.5K & 15.2K
\end{tabular}
\caption{Dataset summary ($K=\times10^3$), (h,n) human- and network-labelled teeth masks.
}
\label{tab:db}
\end{wraptable}

\textbf{Noise, blur, colorization:} $B$ and $R$. We use frame-to-frame (F2F) training \cite{ehret2019model} on static video fragments recorded with camera $C_1$ and apply the trained image denoiser to obtain \textit{denoised frames} and their \textit{noise maps}. The denoised frames are temporally interpolated~\cite{niklaus2017video} 8 times and averaged over a temporal window of $17$ frames to synthesize realistic blur. The noise maps are added to the blurry frames to form the \textit{input video frames} $B$. However, perfect registration of frames between two different modalities $C_1$ and $C_2$ is challenging. Instead, we colorize frames of $C_1$ as per $C_2$ to create the ground truth \textit{output video frames} $R$. Similarly to~\cite{zhang2020rt}, we learn a color mapping (CM) network to predict parameters of 3D functions used to map colors of dental scenes from $C_2$ to $C_1$. In effect, we circumvent local registration errors and obtain exact, pixel-to-pixel spatial correspondence of frames $B$ and $R$.

% The denoised frames of $C_1$ are colorized to create ground truth \textit{output video frames} $R$. Specifically, perfect registration of frames between two different modalities $C_1$ and $C_2$ is challenging. Similarly to~\cite{zhang2020rt}, we learn a color mapping (CM) network to predict parameters of 3D functions 
% (one per each color channel) which are
% used to map colors of dental scenes from $C_2$ to $C_1$.
%to map colors of dental scenes from $C_2$ to $C_1$ via a $8\times8\times8$ color cube. 
%\pko{... we learn a shallow VGG-like color mapping (CM) network to predict parameters of three 3D functions (one per each color channel) which are used to map colors of dental scenes from $C_2$ to $C_1$}. 
% In effect, we circumvent local registration errors and obtain exact, pixel-to-pixel spatial correspondence of frames $B$ and $R$.

% \begin{table}
% \centering
% \begin{tabular}{ c|c|c|c|c  }
% data     & total & train & val  & test \\ \hline 
% videos/frames/ segm (h)  & 409/81K/696   & 300/60K/300   & 29/5.6K/116   & 80/15.5K/320  \vspace{0.25cm} 
% %frames   & 81K   & 60K   & 5.6K & 15.5K   \\
% %segm (h) & 696   & 300   & 116  & 320
% \end{tabular}
% \caption{Dataset summary (K$=\times10^3$), (h) human-labelled teeth masks.}
% \label{tab:db}
% \end{table}

\textbf{Segmentation masks  and homographies:} $M, H$. We manually annotate one frame $R$ of natural teeth in phantom scenes from each training video and four frames of teeth in each validation and test videos. Following \cite{nekrasov2019real} that used a powerful network, an HRNet48~\cite{sun2019deep} pretrained on ImageNet, is fine-tuned on our annotations to automatically segment the teeth in the remaining frames in all three sets. We compute optical flows between consecutive clean frames with RAFT \cite{teed2020raft}. Motion fields are cropped with teeth masks $M_t$ to discard other moving objects, such as the dental bur or the suction tube, as we are interested in stabilizing the videos with respect to the teeth. Subsequently, a partial affine homography $H$ is fitted by RANSAC to the segmented motion field.

% Labels for scale $s>1$ are obtained using inverse relation to Eq.~\ref{eq:output_scale}, i.e. for each $i$, $\task_{i}^s = u_{i}^{-1}(\task_{i}^{s-1})$, where $u_{i}^{-1}$ is a down-sampling and down-scaling for tasks 1,2 and 3, respectively.
%Similarly to (cite), our models optimized within the Imitation Learning paradigm. 
%The resultant dataset consists of 409 sequences of 81086 frames split into training (300 of 59990) validation (29 of 5605) and testing (80 of 15491). A total 696 frames is annotated by humans and the rest in interpolated by HRNet48. Additionally we provide 80687 homography matrices that align consecutive frames. 

%Ubuntu v20.04, NVidia GPU RTX 2080S Turbo, PyTorch v1.10, Torchscript, Kornia v0.6, FP32, FP16.

{\textbf{Setup.}} We train, validate, and test all methods on our dataset (Tab. \ref{tab:db}). In all MOST-Net training runs, we set $\lambda_{1}, \lambda_{2}, \lambda_{3}$ to $2\times10^{-4}$, $5\times10^{-5}$ and 1 for balancing tasks in Eq. \ref{eq:loss}. We train all methods with batch size $16$ and use Adam optimizer with learning rate $1e-4$, decayed to $1e-6$ with cosine annealing. The training frames are augmented by horizontal and vertical flips with $0.5$ probability, random channel perturbations, and color jittering, after~\cite{zhou2019stfan}. All experiments are performed with PyTorch 1.10 (FP32). The inference speed is reported in frames-per-second (FPS) on GPU NVidia RTX 5000.

\begin{figure}[!htbp]
    \centering
    \includegraphics[width=0.95\textwidth]{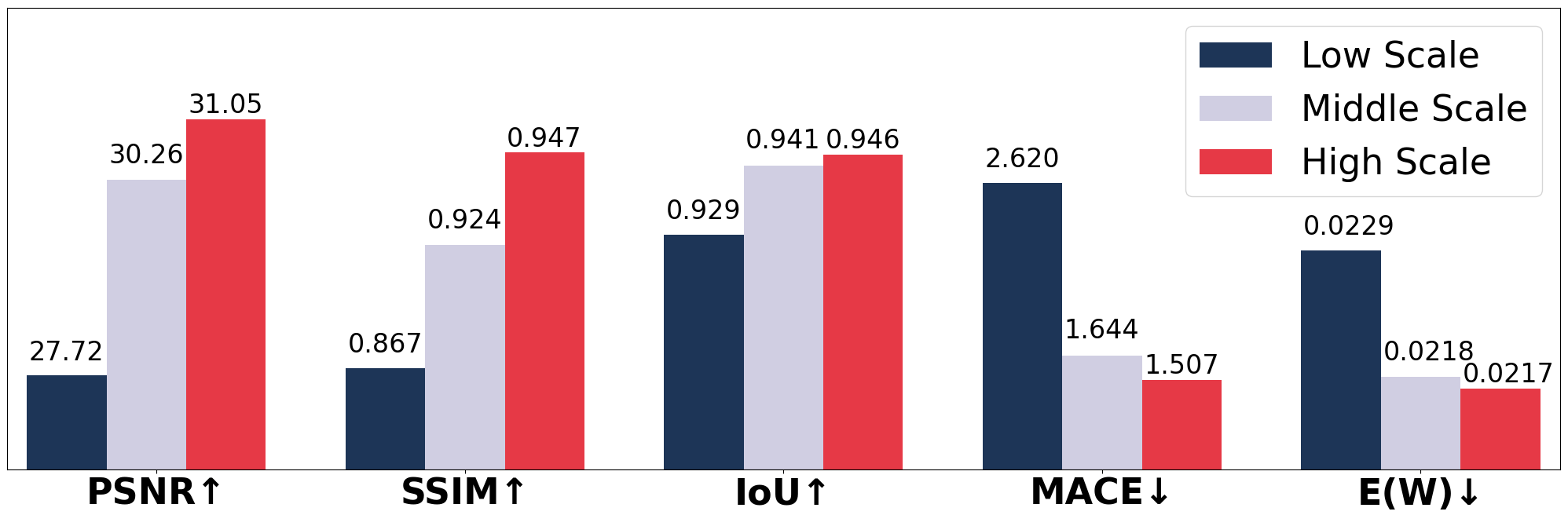}
    \caption{MOST-Net performance improves with output upscaling.}
    \label{fig:multiscale}
\end{figure}

\begin{figure}[!htbp]
    \centering
    \includegraphics[width=0.9\textwidth]{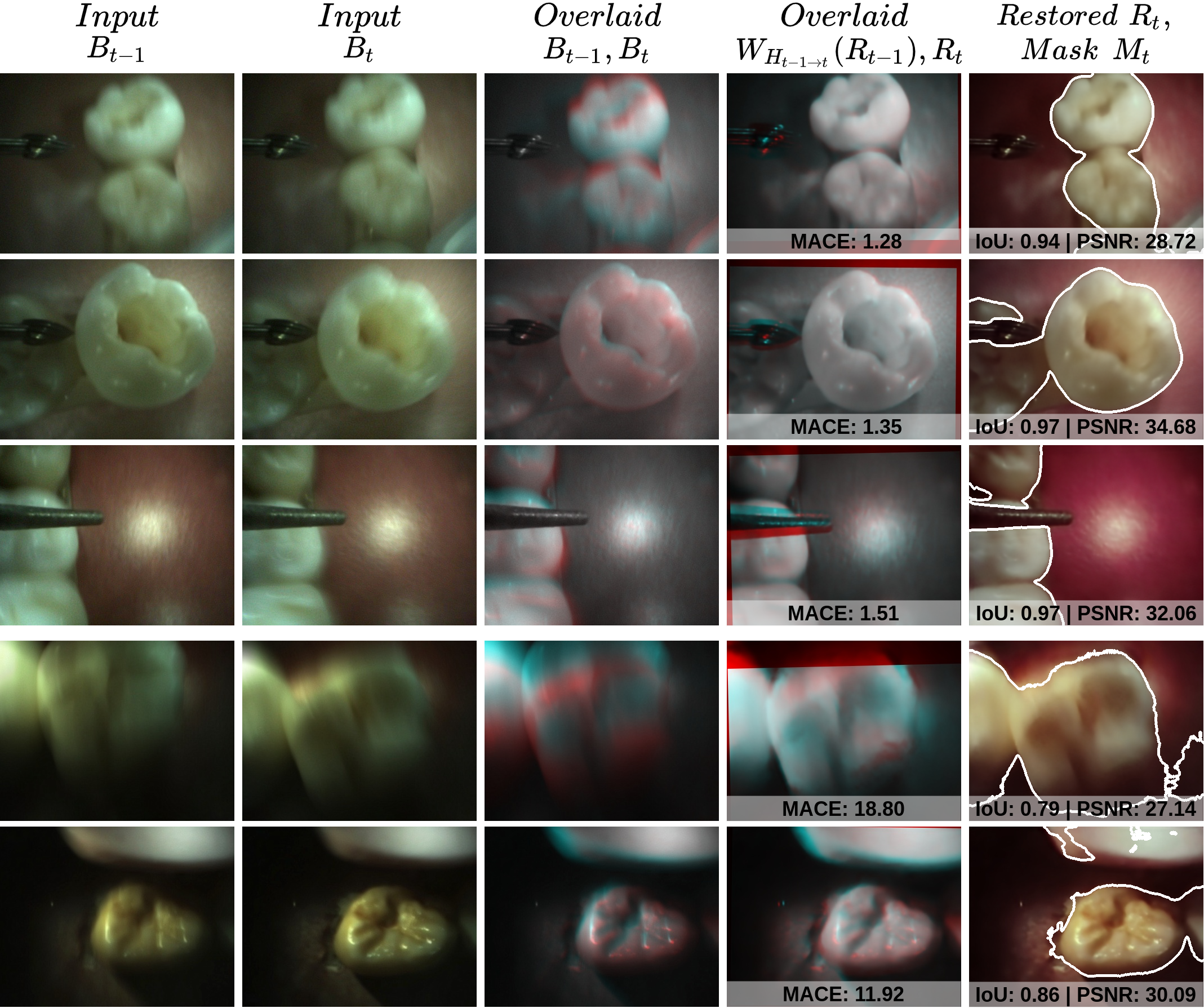}
   % \caption{MOST-Net qualitative results of teeth-specific homography estimation (4th column) and full frame restoration and teeth segmentation (5th column). Failure cases (bottom panel) in registration and segmentation stem from heavy blur, texture paucity, low-light, and tooth-like objects (e.g. suction devices).
   % }
    \caption{Our qualitative results of teeth-specific homography estimation (4th column) and full frame restoration and teeth segmentation (5th column). MOST-Net can denoise video frames and translate pale colors (first and second column) into vivid colors (5th column). Simultaneously, it can deblur and register frames wrt to teeth (4th column). In addition, despite blurry edges in the inputs, MOST-Net produces segmentation masks that align well with teeth contours (rows 1-3). 
    %The 4-5th rows shows the remain challenges, i.e. MOST-Net may fail in case of very heavy blur, or presence of independently moving objects with appearance similar to teeth as in 5th row showing white dental suction tube.
    Failure cases (bottom panel, 4-5th rows) stem from heavy blur (4th row, 
    %texture paucity, low-light, 
    and tooth-like independently moving objects (5th row), such as suction devices.
    }
    \label{fig:qual}
\end{figure}

{\textbf{Diagnostics.}} In Fig. \ref{fig:multiscale} we assess the performance gains across scale levels of MOST-Net. To this end, we upsample all outputs at lower scales to original scale and compare them with ground truth. We observe that MOST-Net performance improves via task-output propagation across scales in all measures. We perform an ablation study of MOST-Net (Tab. \ref{tab:sota}) by reconfiguring our architecture (Fig. \ref{fig:arch}) as follows: (i) NS-no segmentation as auxiliary task, (ii) NE-no connection of encoder features $f_t^s$ with Motion Gated Attention module, (iii) NW-no warping of previous encoder features $f_{t-1}^s$, (iv) NMO-no multi-outputs at scales $s>1$ so that our network has no task interactions between scales. Ablations show that all network configurations lead to $>0.5$dB drops in PSNR and drops in temporal consistency error E(W). Segmentation task and temporal alignment help the video restoration task the most. No multi-task interactions across scales increases MACE error by $>0.6$. The NE ablation improves MACE only slightly at the considerable drop in PSNR. We also find that the IoU remains relatively unaffected by the ablations suggesting room for improving task interactions to aid the figure-ground segmentation task. Qualitative results are shown in Fig.~\ref{fig:qual}.

%Tab. \ref{tab:sota} shows quantitative results and Fig. \ref{fig:qual} illustrates qualitative results.
\begin{table}[!htbp]
    \centering
    \begin{tabular}{ |l | c | c | c | c | c | c | c |}
\hline
         Methods         & PSNR $\uparrow$& SSIM $\uparrow$ & MACE $\downarrow$ & IoU $\uparrow$ & E(W) $\downarrow$  & \#P(M) & FPS \\
\hline
\hline
         MIMO-UNet~\cite{cho2021rethinking}         &   26.66 & 0.916             &      -    &   -         &   0.0278  & 5.3 &  8.4    \\
         ESTRNN~\cite{zhong2020efficient}          &   30.72 & 0.943             &      -    &   -         &   0.0229  & 2.3 & 68.5 \\ %67.2    \\

\hline
         MHN~\cite{le2020deep} &    -  & -                    &  1.347    &   -        &   -        & 6.2 & 89.8 \\%84.9%    \\
        %  Le et al. (data red5X) &  -            &  7.017    &   -              &  &  &              %  \\
        %  Le et al. (data red5X, on GT) &  -     &  6.042    &   -              &  &  &              %  \\
%         MHN (GT)      &  -                        & 0.608     &   -        &        -   &     &         \\
\hline 
         DeepLabv3+(DL)~\cite{chen2018encoder}       &         - & -      &     -     &  0.968 & -   & 26.7 & 108.2 \\ %49.0%  \\
         UNet++~\cite{zhou2018unet++}           &         -  & -     &     -     &  0.969 &   -  & 50.0 &  38.9 \\%33.4%  \\
\hline
        %  ESTRNN + DL (GT) &   30.61/0.942       &     -      &   0.911&     -      & 29.0  & 46.1  & 27.1   \\
        %  ESTRNN + DL (CR) &   30.61/0.942       &     -      &   0.444 &   -       & 29.0  & 46.1  & 27.1   \\
%         ESTRNN+DL &   30.61 / 0.942       &     -      &   0.959  &    -     & 29.0  & 27.1   \\

        %  ESTRNN + Le (GT) &   30.61/0.942       &     2.061 &   -  &    -         & 8.5  & 39.5  & 37.6   \\
        %  ESTRNN + Le (CR) &   30.61/0.942       &     1.539 &   -  &     -        & 8.5  & 39.5  & 37.6   \\
%         ESTRNN+MHN&   30.61 / 0.942       &     1.377 &   -  &    -         & 8.5   & 37.6   \\
         ESTRNN+MHN+DL&   30.72 & 0.943       & \textbf{1.368}  &  \textbf{0.967} &    0.0229       & 35.2  & \textbf{28.6}   \\
\hline
\hline
        MOST-Net-NS     &     30.21 & 0.939    &  1.426    &  - &   0.0223  & 9.7 &  19.0 \\
        MOST-Net-NE     &     30.22 & 0.941    &  1.423    &  0.946  & 0.0221 & 9.8 &  19.2 \\
        MOST-Net-NW     &     30.37 & 0.943    &  1.456    &  0.952  &  0.0221 & 9.8 &  19.3 \\
        MOST-Net-NMO    &     30.48 & 0.940    &  2.155    &  0.946 & 0.0227  & 8.5 & 19.1 \\
        \hline 
        MOST-Net        &     \textbf{31.05} & \textbf{0.947}    &  1.507    &  0.946  & \textbf{0.0217}  & \textbf{9.8} & 19.3\\
        \hline

        % MOST-Net-RB...        &   30.19   & 0.940    &  1.634   & 0.930  &   & 8.7 & 27.3\\
        
        % MOST-Net-RB-DWA       &  30.49    & 0.940    &  1.637   & 0.939  &   & 8.7 & 27.3\\

        % MOST-Net-RDB          &   30.49   &  0.944   &  1.667   & 0.919  &   & 6.5 & 25.1\\

        % MOST-Net-RDB-DWA-8 DOF &           &          &           &    &   & 6.5 & 25.1\\

        \hline
    \end{tabular}
    \caption{STL and MTL benchmarks (top panel) and MOST-Net (bottom panel). Best results of MOST-Net wrt ESTRNN+MHN+DL are in bold.}
    %\caption{The proposed architecture is tailored bearing a maximally synergic setup whereas gradients from tasks communicate optimally. Indeed, adding new tasks in the proposed architecture does not impact the performance of their single-task counterparts.}
    \label{tab:sota}
\end{table}

%\aj{The 4-5th rows shows the remain challenges, i.e. MOST-Net may fail in case of very heavy blur, or presence of independently moving objects with appearance similar to teeth as in 5th row showing white dental suction tube. }

%We refer to the supplementary material for comparisons with other methods.

%We compare MOST-Net with state-of-the-art single task restoration baselines ESTRNN~\cite{zhong2020efficient} and MIMO-UNet~\cite{cho2021rethinking} in terms of PSNR and SSIM. MOST-Net outperforms both approaches.

%Indeed, progressive temporal alignment as manifested via warping, segmentation and multi-scales boost restoration metrics. 

{\textbf{Quantitative results.}} We compare MOST-Net with single task state-of-the-art methods for restoration, homography estimation, and binary segmentation in Tab.~\ref{tab:sota}. MOST-Net outperforms video restoration baseline ESTRNN~\cite{zhong2020efficient} and image restoration MIMO-UNet~\cite{cho2021rethinking} in PSNR by $>0.3$dB and $>4.3$dB, respectively. We posit the low PSNR performance of MIMO-UNet stems from its single frame input that negatively affects its colorization abilities and also leads to high temporal consistency error E(W)~\cite{lai2018learning}. ESTRNN also introduces observable flickering artifacts expressed by higher E(W) than MOST-Net. MACE error in homography estimation is slightly higher for our method than for MHN~\cite{le2020deep} but MOST-Net has potential for improvement due to its multi-tasking approach. Notably, MHN has a significantly lower error of MACE=$0.6$ when it is trained and tested on \textit{ground truth} videos, which have vivid colors and no noise and blur. This suggests that video restoration task is necessary to improve homography estimation task. Subsequently, we evaluate our method wrt DeepLabv3+~\cite{chen2018encoder} with ResNet50 encoder and wrt UNet++~\cite{zhou2018unet++} for teeth segmentation task using intersection-over-union (IoU) criterion. MOST-Net achieves comparable results with these benchmarks with several times less parameters (\#P(M)). Though both methods have several times higher FPS, MOST-Net addresses three tasks instead of a single task and still achieves near real-time efficiency. Finally, we compare our multi-task MOST-Net with single task methods ESTRNN+MHN+DL that are forked, with MHN and DL as heads. ESTRNN restores videos from the training set and MHN and DeepLabv3+ (DL) are trained and tested on the restored frames. The pipeline runs at 28.6 FPS but requires $\times3.6$ more model parameters than our network. Moreover, MOST-Net achieves higher PSNR, SSIM, and E(W) results than the forked pipeline on the video restoration task, having comparable MACE error and IoU scores while running near real time at 19.3 FPS or \textbf{21.3 FPS} (TorchScript-ed).

%\section{Conclusions} 
{\textbf{Conclusions.}} We proposed MOST-Net, a novel deep network for video processing that models task interactions across scales. MOST-Net jointly addressed the tasks of video restoration, teeth segmentation, and homography-based motion estimation. The study demonstrated the applicability of the network in computer-aided dental intervention on the publicly released \textit{Vident-lab} video dataset of natural teeth in phantom scenes.

\bibliographystyle{splncs04}
\bibliography{mybibliography}

\begin{thebibliography}{10}
\providecommand{\url}[1]{\texttt{#1}}
\providecommand{\urlprefix}{URL }
\providecommand{\doi}[1]{https://doi.org/#1}

\bibitem{bradley2021cinematic}
Bradley, A., Klivington, J., Triscari, J., van~der Merwe, R.: Cinematic-l1
  video stabilization with a log-homography model. In: Proceedings of the
  IEEE/CVF Winter Conference on Applications of Computer Vision. pp. 1041--1049
  (2021)

\bibitem{charbonnier1994two}
Charbonnier, P., Blanc-Feraud, L., Aubert, G., Barlaud, M.: Two deterministic
  half-quadratic regularization algorithms for computed imaging. In:
  Proceedings of 1st International Conference on Image Processing. vol.~2, pp.
  168--172. IEEE (1994)

\bibitem{chen2018encoder}
Chen, L.C., Zhu, Y., Papandreou, G., Schroff, F., Adam, H.: Encoder-decoder
  with atrous separable convolution for semantic image segmentation. In:
  Proceedings of the European conference on computer vision (ECCV). pp.
  801--818 (2018)

\bibitem{cho2021rethinking}
Cho, S.J., Ji, S.W., Hong, J.P., Jung, S.W., Ko, S.J.: Rethinking
  coarse-to-fine approach in single image deblurring. In: Proceedings of the
  IEEE/CVF International Conference on Computer Vision. pp. 4641--4650 (2021)

\bibitem{detone2016deep}
DeTone, D., Malisiewicz, T., Rabinovich, A.: Deep image homography estimation.
  arXiv preprint arXiv:1606.03798  (2016)

\bibitem{ehret2019model}
Ehret, T., Davy, A., Morel, J.M., Facciolo, G., Arias, P.: Model-blind video
  denoising via frame-to-frame training. In: Proceedings of the IEEE/CVF
  Conference on Computer Vision and Pattern Recognition. pp. 11369--11378
  (2019)

\bibitem{jung2021multi}
Jung, H., Kim, Y., Jang, H., Ha, N., Sohn, K.: Multi-task learning framework
  for motion estimation and dynamic scene deblurring. IEEE Transactions on
  Image Processing  \textbf{30},  8170--8183 (2021)

\bibitem{katsaros2021concurrent}
Katsaros, E., Ostrowski, P.K., Wesierski, D., Jezierska, A.: Concurrent video
  denoising and deblurring for dynamic scenes. IEEE Access  \textbf{9},
  157437--157446 (2021)

\bibitem{kokkinos2017ubernet}
Kokkinos, I.: Ubernet: Training a universal convolutional neural network for
  low-, mid-, and high-level vision using diverse datasets and limited memory.
  In: Proceedings of the IEEE conference on computer vision and pattern
  recognition. pp. 6129--6138 (2017)

\bibitem{kuhnisch2021caries}
K{\"u}hnisch, J., Meyer, O., Hesenius, M., Hickel, R., Gruhn, V.: Caries
  detection on intraoral images using artificial intelligence. Journal of
  dental research  (2021)

\bibitem{lai2018learning}
Lai, W.S., Huang, J.B., Wang, O., Shechtman, E., Yumer, E., Yang, M.H.:
  Learning blind video temporal consistency. In: Proceedings of the European
  conference on computer vision (ECCV). pp. 170--185 (2018)

\bibitem{le2020deep}
Le, H., Liu, F., Zhang, S., Agarwala, A.: Deep homography estimation for
  dynamic scenes. In: Proceedings of the IEEE/CVF Conference on Computer Vision
  and Pattern Recognition. pp. 7652--7661 (2020)

\bibitem{low2018magnification}
Low, J.F., Dom, T.N.M., Baharin, S.A.: Magnification in endodontics: A review
  of its application and acceptance among dental practitioners. European
  journal of dentistry  \textbf{12}(04),  610--616 (2018)

\bibitem{mao2021deep}
Mao, X., Liu, Y., Shen, W., Li, Q., Wang, Y.: Deep residual fourier
  transformation for single image deblurring. arXiv preprint arXiv:2111.11745
  (2021)

\bibitem{marstal2016simpleelastix}
Marstal, K., Berendsen, F., Staring, M., Klein, S.: Simpleelastix: A
  user-friendly, multi-lingual library for medical image registration. In: IEEE
  CVPR Workshops. pp. 134--142 (2016)

\bibitem{misra2016cross}
Misra, I., Shrivastava, A., Gupta, A., Hebert, M.: Cross-stitch networks for
  multi-task learning. In: Proceedings of the IEEE conference on computer
  vision and pattern recognition. pp. 3994--4003 (2016)

\bibitem{murphy2012machine}
Murphy, K.P.: Machine learning: a probabilistic perspective. MIT press (2012)

\bibitem{nekrasov2019real}
Nekrasov, V., Dharmasiri, T., Spek, A., Drummond, T., Shen, C., Reid, I.:
  Real-time joint semantic segmentation and depth estimation using asymmetric
  annotations. In: 2019 International Conference on Robotics and Automation
  (ICRA). pp. 7101--7107. IEEE (2019)

\bibitem{niklaus2017video}
Niklaus, S., Mai, L., Liu, F.: Video frame interpolation via adaptive separable
  convolution. In: Proceedings of the IEEE International Conference on Computer
  Vision. pp. 261--270 (2017)

\bibitem{rashid2022hybrid}
Rashid, U., Javid, A., Khan, A.R., Liu, L., Ahmed, A., Khalid, O., Saleem, K.,
  Meraj, S., Iqbal, U., Nawaz, R.: A hybrid mask rcnn-based tool to localize
  dental cavities from real-time mixed photographic images. PeerJ Computer
  Science  (2022)

\bibitem{ronneberger2015u}
Ronneberger, O., Fischer, P., Brox, T.: U-net: Convolutional networks for
  biomedical image segmentation. In: International Conference on Medical image
  computing and computer-assisted intervention. pp. 234--241. Springer (2015)

\bibitem{sun2019deep}
Sun, K., Xiao, B., Liu, D., Wang, J.: Deep high-resolution representation
  learning for human pose estimation. In: Proceedings of the IEEE/CVF
  Conference on Computer Vision and Pattern Recognition. pp. 5693--5703 (2019)

\bibitem{teed2020raft}
Teed, Z., Deng, J.: Raft: Recurrent all-pairs field transforms for optical
  flow. In: European conference on computer vision. pp. 402--419. Springer
  (2020)

\bibitem{vandenhende2020mti}
Vandenhende, S., Georgoulis, S., Gool, L.V.: Mti-net: Multi-scale task
  interaction networks for multi-task learning. In: European Conference on
  Computer Vision. pp. 527--543. Springer (2020)

\bibitem{vandenhende2021multi}
Vandenhende, S., Georgoulis, S., Van~Gansbeke, W., Proesmans, M., Dai, D.,
  Van~Gool, L.: Multi-task learning for dense prediction tasks: A survey. IEEE
  transactions on pattern analysis and machine intelligence  (2021)

\bibitem{wang2019edvr}
Wang, X., Chan, K.C., Yu, K., Dong, C., Change~Loy, C.: Edvr: Video restoration
  with enhanced deformable convolutional networks. In: CVPR Workshops (2019)

\bibitem{xu2018pad}
Xu, D., Ouyang, W., Wang, X., Sebe, N.: Pad-net: Multi-tasks guided
  prediction-and-distillation network for simultaneous depth estimation and
  scene parsing. In: CVPR. pp. 675--684 (2018)

\bibitem{zhang2020rt}
Zhang, M., Gao, Q., Wang, J., Turbell, H., Zhao, D., Yu, J., Lu, Y.:
  {RT-VENet}: A convolutional network for real-time video enhancement. In:
  Proceedings of the 28th ACM International Conference on Multimedia. pp.
  4088--4097 (2020)

\bibitem{zhang2019pattern}
Zhang, Z., Cui, Z., Xu, C., Yan, Y., Sebe, N., Yang, J.: Pattern-affinitive
  propagation across depth, surface normal and semantic segmentation. In: CVPR.
  pp. 4106--4115 (2019)

\bibitem{zhong2020efficient}
Zhong, Z., Gao, Y., Zheng, Y., Zheng, B.: Efficient spatio-temporal recurrent
  neural network for video deblurring. In: European Conference on Computer
  Vision. pp. 191--207. Springer (2020)

\bibitem{zhou2019stfan}
Zhou, S., Zhang, J., Pan, J., Xie, H., Zuo, W., Ren, J.: Spatio-temporal filter
  adaptive network for video deblurring. In: Proceedings of the IEEE
  International Conference on Computer Vision (2019)

\bibitem{zhou2018unet++}
Zhou, Z., Rahman~Siddiquee, M.M., Tajbakhsh, N., Liang, J.: Unet++: A nested
  u-net architecture for medical image segmentation. In: Deep learning in
  medical image analysis and multimodal learning for clinical decision support,
  pp. 3--11. Springer (2018)

\bibitem{zhu2020tooth}
Zhu, G., Piao, Z., Kim, S.C.: Tooth detection and segmentation with mask r-cnn.
  In: 2020 International Conference on Artificial Intelligence in Information
  and Communication (ICAIIC). pp. 070--072. IEEE (2020)

\end{thebibliography}

\end{document}